\documentclass[runningheads]{llncs}
\usepackage[T1]{fontenc}
\usepackage{graphicx}
\usepackage[utf8]{inputenc}
\usepackage{amsmath,epsfig}
\usepackage[ruled,linesnumbered]{algorithm2e}
\usepackage{bbding}
\usepackage{float}
\usepackage{pifont}
\usepackage{subfig}
\usepackage{url}
\usepackage{booktabs}
\usepackage{amssymb}
\usepackage{bbding}
\usepackage{pifont}
\usepackage{wasysym}
\usepackage{utfsym}
\usepackage{fontawesome}
\usepackage{array}
\usepackage{cite} 

\usepackage[final]{changes} 
\definechangesauthor[name={Author1}, color=red]{Change}

\begin{document}
\title{Learning to Generalize Unseen Domains via Multi-Source Meta Learning for Text Classification}
%
%
\author{Yuxuan Hu\inst{1,2}\orcidID{0009-0005-8571-118X} \and
Chenwei Zhang\inst{1,2}\orcidID{0009-0000-6698-0652} \and
Min Yang\inst{4}\orcidID{0000-0001-7345-5071} \and
Xiaodan Liang \inst{1}\orcidID{0000-0003-3213-3062} \and
Chengming Li\inst{2,3}\orcidID{0000-0002-4592-3875} \and
Xiping Hu\inst{2,3}\orcidID{0000-0002-4952-699X}}
\authorrunning{Yuxuan Hu, et al.}
%
\institute{
Shenzhen MSU-BIT University, Shenzhen, Guangdong, China\\
\email{\{licm,huxp\}@smbu.edu.cn} \and
Sun Yat-Sen University, Shenzhen, Guangdong, China\\ 
\email{\{huyx55,zhangchw7\}@mail2.sysu.edu.cn, liangxd9@mail.sysu.edu.cn}\and
Guangdong-Hong Kong-Macao Joint Laboratory for Emotional Intelligence and Pervasive Computing, Shenzhen MSU-BIT University, Shenzhen, Guangdong, China. \and
SIAT, Chinese Academy of Sciences, Shenzhen, Guangdong, China\\
\email{min.yang@siat.ac.cn}}
\maketitle              
\begin{abstract}
With the rapid development of deep learning methods, there have been many breakthroughs in the field of text classification. Models developed for this task have \replaced{achieved}{been shown to achieve}high accuracy. However, most of these models are trained using labeled data from seen domains. It is difficult for these models to maintain high accuracy in a new challenging unseen domain, which is directly related to the generalization of the model. In this paper, we study the multi-source Domain Generalization \replaced{for}{of} text classification and propose a framework to use multiple seen domains to train a model that can achieve high accuracy in an unseen domain. Specifically, we propose a multi-source meta-learning Domain Generalization framework to simulate the process of model generalization to an unseen domain, so as to extract sufficient domain-related features. We \replaced{introduce}{introduced} a memory mechanism to store domain-specific features, which coordinate with the meta-learning framework. Besides, we adopt \replaced{a}{the} novel "jury" mechanism that enables the model to learn sufficient domain-invariant \deleted{classification} features. Experiments demonstrate that our meta-learning framework can effectively enhance the ability of the model to generalize to an unseen domain and can outperform the state-of-the-art methods on multi-source text classification datasets.

\end{abstract}
\section{Introduction}
The text classification of social media is crucial not only for conducting surveys among traditional consumers and companies to gather opinions on respective products or services\added{,} but also plays a significant role in national security and public opinion analysis \cite{blitzer2007biographies,zubiaga2016analysing}. While recent deep learning models of text classification  \cite{yang2019xlnet,liu2019roberta,sanh2019distilbert} demonstrate efficacy \replaced{in}{on} a seen domain (i.e., a domain with labeled data), most of them do not perform well in an unseen domain (i.e., a domain only with unlabeled data). However, in real life, text classification is inevitably used in unseen domains. Text classification can be \replaced{considered}{regarded} as a domain-dependent task, because a sentence may convey different meanings in various domains. For instance, the term "short" in the context of "short service time" in an electronic review is construed as negative, whereas in a restaurant review, "short" in the context of "short waiting time" is considered positive. 

\added{Domain generalization(DG) is a type of transfer learning task. A similar task is Domain Adaptation, which allows access to both source domain and target domain data. Unlike Domain Adaptation, Domain Generalization only permits access to data from the visible domains.} The goal of the Domain generalization approach is to \replaced{address}{solve} this problem by training a well-generalized model only with labeled data from one or more seen source domains and testing on an unseen domain. Although there has been considerable research on DG in the field of image classification, there have been few studies \replaced{address}{solve} in DG of text classification. Most studies in DG of text classification are based on Mixture of Experts (MoE) \cite{guo2018multi,li2018s}. This approach involves training domain-specific experts and a domain-shared expert independently, followed by their aggregation using a score function. However, the effectiveness of these methods is constrained. \replaced{In contrast}{Contrastingly}, in real-world scenarios, individuals exhibit a natural ability to swiftly adapt to texts in unknown domains.  We posit that the differentiating factor between humans and machines lies in humans' capacity to autonomously categorize domain knowledge into domain-specific and domain-invariant categories and form semantic memory. This ability enables humans to enhance their generalization capacity \replaced{using}{with the aid of} prior knowledge. Inspired by this, we contend that storing domain-specific knowledge and domain-invariant knowledge \deleted{between domains} can enhance the DG capabilities of the model.

In this paper, we propose a \textbf{M}ulti-source \textbf{M}eta-learning framework relying on \added{a} "\textbf{J}ury" mechanism and \textbf{M}emory module (\textbf{MMJM}) to facilitate the learning of both \deleted{learn} domain-invariant and domain-specific features. Specifically, we introduce a meta-learning framework \cite{zhao2021learning} based on the multi-source DG, which simulates \replaced{how the model generalizes}{the process of model generalization} to an unseen domain. We suppose that the meta-learning approach aids the model in differentiating domains and \replaced{learning}{studying} the way of classification in an unseen domain by enhancing its ability to capture domain-related features. Additionally, we incorporate a memory mechanism \cite{zhao2021learning} to more effectively capture domain-specific features, leveraging domain information comprehensively while mitigating the risk of unstable optimization. Furthermore, we introduce an innovative "Jury" mechanism \cite{chen2021style} to exploit domain-invariant features. This mechanism promotes \deleted{the} features from the same class to be closer and features from different classes to be further away.

Our contributions are summarized as follows:
\begin{itemize}
\item We propose a novel multi-source meta-learning framework with the memory and "jury" mechanism, which simulates \replaced{how the model generalizes}{the process of model generalization} to an unseen domain.
\item We are the first to introduce a memory mechanism into DG \replaced{for}{in} text classification, aimed at learning both domain-specific and domain-invariant features.
\item We demonstrate the effectiveness of our proposed framework through extensive experiments and detailed analyses.
\end{itemize}

\section{Related Work}
\textbf{Domain Generalization (DG).} In the DG method, the model can access one or more source domains only with labeled data and needs to be tested in the target domain with unlabeled data. Most \deleted{of the} previous research \replaced{focuses on extracting}{work on extract} domain-invariant representation. The first widely explored method is domain adversarial training. Many studies \cite{guo2018multi,li2018s, chen2018multinomial,wright2020transformer} use this to reduce the divergence between domains. Another method is based on a mixture of experts (MoE). For example, the work \cite{guo2018multi} \replaced{propose}{proposed} a set of parallel domain-specific experts to get \replaced{multiple}{many} domain-related classification results and \replaced{use}{then used} a distance metric component to compute the mix score to choose the expert. \added{Recently, with the impressive advancements in large language models, many have demonstrated excellent generalization capabilities. Models like ChatGPT (OpenAI), ChatGLM~\cite{zeng2022glm, du2022glm}, Llama~\cite{touvron2023llama}, Mixtral~\cite{jiang2024mixtral}, and others have shown outstanding abilities in domain generalization. These models exhibit excellent classification performance across various domains, demonstrating remarkable domain generalization capabilities.}

\textbf{Meta-Learning.} The concept of meta-learning is “learning to learn” \cite{thrun1998learning}. Its main idea is to divide the training stage into \added{a} meta-train stage and \added{a} meta-test stage, using multi-step gradient descent to learn a good initialization. MLDG \cite{li2018learning} \replaced{is}{was} the first to incorporate meta-learning into DG, innovatively transforming the meta-train and meta-test process to simulate domain shift situations. \replaced{This development catalyzes}{Subsequently, this development catalyzed} the integration of various meta-learning methods into DG. For example, \cite{balaji2018metareg} \replaced{suggests introducing a regularization function for meta-learning}{suggested using a regularization function to encode DG}, which is called meta regularizer (MetaReg). What is relevant to us is \replaced{\cite{zhao2021learning}, which introduces}{that \cite{zhao2021learning} introduced} a meta-learning framework with a memory module in DG for the Re-Identification task. Different from previous methods, we \replaced{are the first to introduce the meta-learning method into DG for text classification, combining it with a memory module and a "jury" module.}{first introduce the meta-learning method into DG in text classification. We combined meta-learning with a memory module and a "jury" module.} 

\textbf{Contrastive Learning.} 
Contrastive Learning \cite{hadsell2006dimensionality} is commonly employed to learn the general features by training models to identify similarities and differences among data points. The representative learning style \cite{oord2018representation} is to make an anchor closer to a "positive" sample and further from many "negative" samples in the representation space. The work \cite{wu2021esimcse} first introduces contrastive learning into text classification. Then, \cite{chen2021style} \replaced{introduces}{introduced} a "jury" mechanism, which can learn domain-invariant features with a memory module. Recently, \cite{tan2022domain} \replaced{introduces}{introduced} supervised contrastive learning into DG in text classification to help "learn an ideal joint hypothesis of the source domains". Inspired by these methods, we introduce the "jury" mechanism into \deleted{the}DG \replaced{for}{in} text classification.

\section{Method}
\subsection{Problem Definition}
The goal of the multi-source DG is to train a model \replaced{using multiple seen source domains}{from multi-seen domains} with labeled data, ensuring its effective performance in previously unseen target domains. In this paper, we consider $D$ \added{seen} source domains $D_{S}=\{D_{S}^d\}_{d=1}^D$ \added{as the training set} and only one unseen target domain $D_{T}$ \added{as the test set}. The data of each source domain is defined as: $D_{S}^d=\{(x_{d,i},y_{d,i})\}_{i=1}^{N_d}$, where $N_d$ \replaced{is the number of samples in the}{is the sample number of the} $d$-th source domain; $x_{d,i}$ is a sample \replaced{from}{of} the $d$-th source domain; $y_{d,i}$ is the label of $x_{d,i}$, where $y_{d,i} = \{ c\}_{c=1}^{N_C}$, $N_C$ is the number of the class.

\subsection{Meta-Learning Framework}
\begin{figure} [t!]
	\centering
	\subfloat[The Overall Meta-leaning Framework. 
	\label{Meta-leaning Framework}]{
		\includegraphics[scale=0.4]{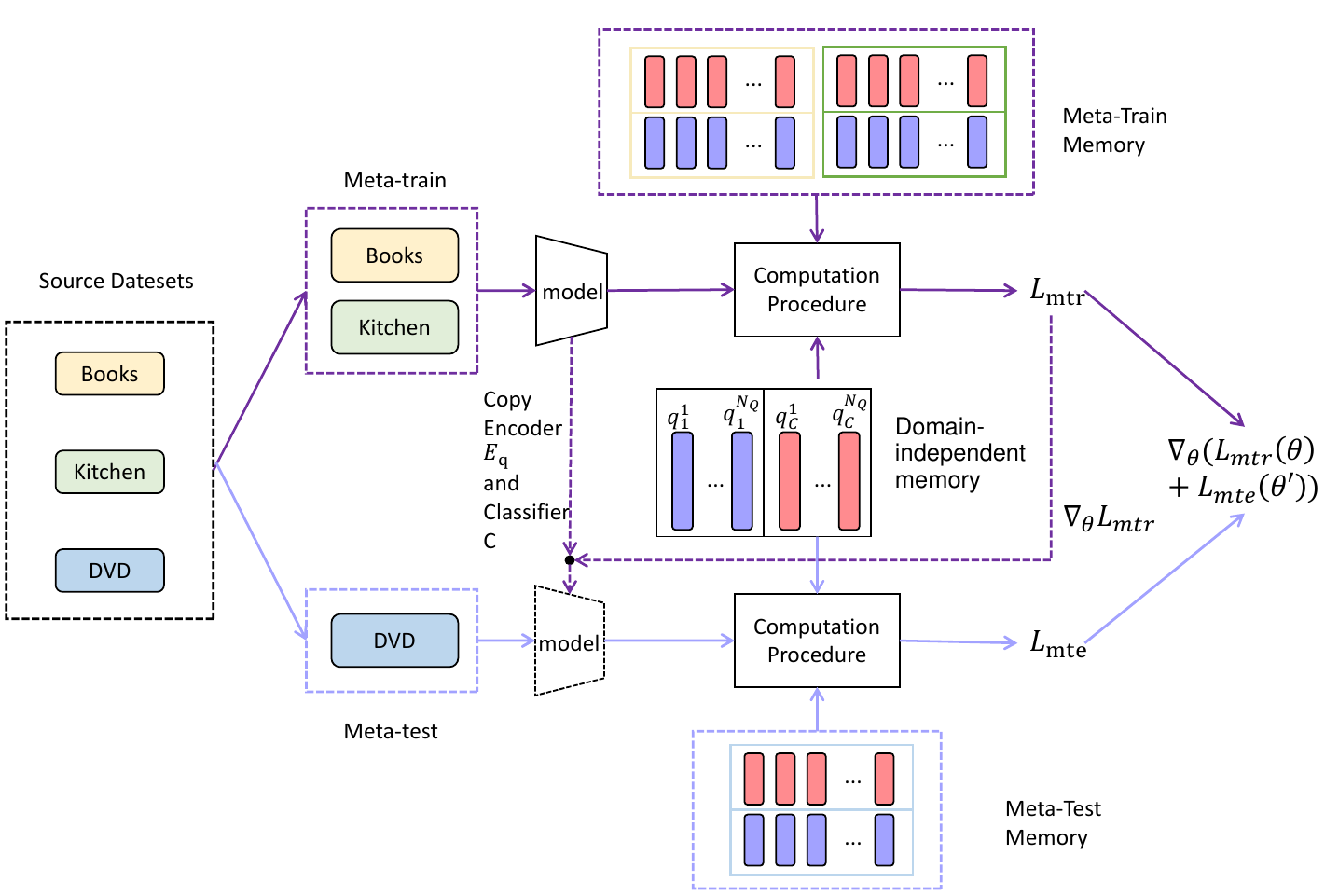}}
	\\
	\subfloat[The Detailed Computation Procedure\label{Computation Procedure}]{
		\includegraphics[scale=0.35]{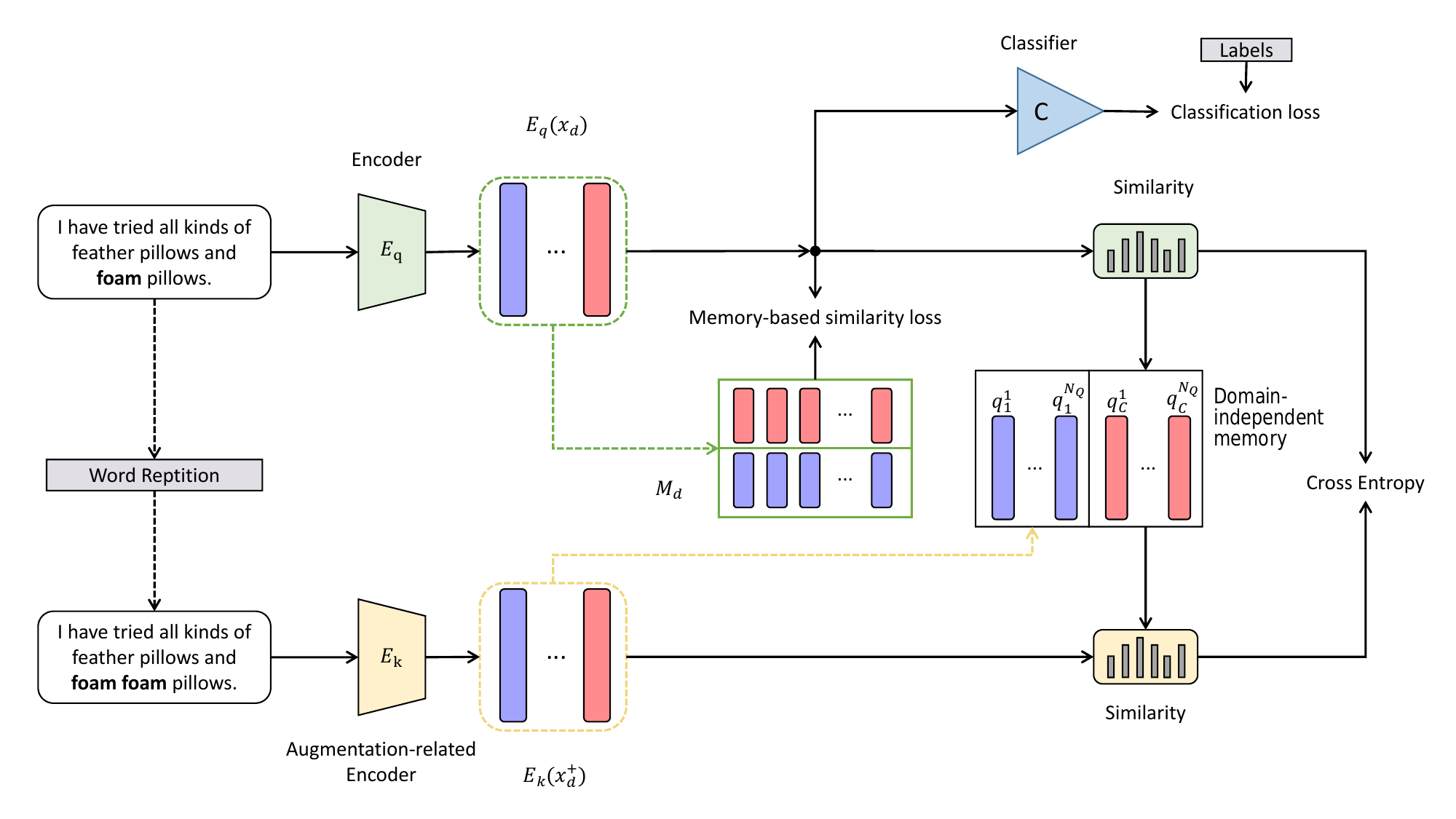} }
	\caption{The proposed framework MMJM. Fig.\textcolor{red}{\ref{Meta-leaning Framework}} \replaced{shows}{is} the overall meta-learning framework and the Fig.\textcolor{red}{\ref{Computation Procedure}} \replaced{shows}{is} the detailed computation procedure of \added{the} meta-train and meta-test stages.}
	\label{model} 
\end{figure}

Following \cite{zhao2021learning}, we introduce meta-learning to simulate \replaced{how the model generalizes}{the process of model generalization} to an unseen domain. We \deleted{try to} divide the training stage into meta-train and meta-test stages. At the beginning of each training epoch, we randomly select one domain's data as the meta-test dataset, and the \replaced{remaining}{rest} $D-1$ domains' data as the meta-train datasets. Our model is shown in \deleted{the} Fig.\textcolor{red}{\ref{model}}.
\setlength{\belowcaptionskip}{-15pt}

We update the model with meta-train and meta-test stages together. At first, we copy the original model. During the meta-train stage, we calculate the meta-train loss $L_{mtr}$ with the original model. This loss is composed of three parts\replaced{: }{. The first is }the classification loss $L_{Class}$ calculated by the classifier $C$\replaced{,}{. The second is} the similarity loss $L_{Mem}$\deleted{, which is} calculated by comparing the features extracted by the encoder with those stored in the memory module\replaced{, and}{. Lastly, the third component is} the $L_{Jury}$\deleted{,} calculated by the "jury" mechanism. Then\added{,} we update the copied model with the meta-train loss. During the meta-test stage, meta-test loss $L_{mte}$ is calculated in a similar way as \added{in} the meta-train stage. Finally, we update the original model with both the meta-train and meta-test \replaced{losses}{loss}. Therefore, the model parameters are updated through the combined meta-train stage and meta-test \replaced{stages}{stage}. The final model update formula is shown in Eq\added{u}.\textcolor{red}{\ref{eqn1}}:
\begin{align}\label{eqn1}
    \mathop{\arg\min}\limits_{\theta_{q,C}}L_{mtr}(\theta_{q,C})+L_{mte}(Adam(\nabla_{\theta_{q,C}}L_{mtr}(\theta_{q,C}),\alpha))
\end{align}
where $\theta_{q,C}$ denotes the parameters of the encoder $E_q$ and the classifier $C$;  $L_{mtr}$ is the meta-train loss; $L_{mte}$ is the meta-test loss; Adam is an optimizer; \added{and} $\alpha$ is the inner loop learning rate.

\subsection{Memory Module}
The memory module is used in conjunction with meta-learning algorithms to store domain-specific features \replaced{for}{of} each domain. We maintain a memory module for each domain\added{,denoted as} $M = \{M_{d}\}_{d=1}^D$, containing features \replaced{for}{of} each class of that domain, where $D$ is the number of source domains. \replaced{Each domain's memory module contains}{The memory module of each domain containing} $N_C$ feature slots $M_{d} = \{M_{d}[c]\}_{c=1}^{N_C}$, where $N_C$ is the number of classes. The dimension of a slot is equal to the dimension of the encoder $E_q$. We calculate the memory-based similarity loss \replaced{using}{with} classification features encoded by the encoder $E_q$ and \replaced{the slot}{slots} $M_{d}[c]$ of each memory.

\textbf{Initialization. }
We initialize all memory modules before the training stage. During initialization, we sequentially initialize all slots in every memory \added{module }of all domains in the same way. Specifically, we \deleted{sequentially} use the pre-trained encoder to \added{sequentially} extract the features of all samples from a class in a source domain. We then \deleted{need to} initialize the corresponding slot with a domain-specific feature. \replaced{We}{Here, we} calculate the mean of the features of each class and \replaced{use}{take} it as the initialization value for each slot.

\textbf{Update. }
After \deleted{the end of} each iteration, \replaced{we}{We} update the slots of each domain sequentially. All slots in all domains are updated in the same way. \replaced{We}{The way we} update a slot \replaced{by encoding the}{is that we encode} features of that class in the current iteration and \replaced{using}{use} a momentum method to update \deleted{them on }the corresponding class slot, as \replaced{shown in Equ}{Eq}.\textcolor{red}{\ref{eqn2}}: 
\begin{align}\label{eqn2}
    M_d[c] = m \cdot M_d[c] + (1-m) \cdot E[c]
\end{align}
where $M_d[c]$ is the slot $c$ of the domain memory $d$; momentum $m$ is a momentum parameter; $E[c]$ consists of all the features of the $c$ class\added{, defined as}: $E[c]=\frac{1}{n}\sum_{i=1}^nE(x_{d,i})$, \added{where }$n$ is the number of samples of the $c$ class in \added{the} $d$ domain \replaced{for}{of} that iteration.

\textbf{Memory-based Similarity Loss. }
We obtain the memory-based similarity loss by calculating the similarity score between \added{the} features encoded by the encoder $E_q$ and the slots in the memory module. We calculate the similarity between \added{the} feature $E(x_{d,i})$ and the \added{corresponding } slot in memory, and normalize \replaced{these values}{them} with softmax. The calculation method is shown in Eq\added{u}.\textcolor{red}{\ref{eqn3}}:
\begin{align}\label{eqn3}
	L_{Mem}=-log \frac{exp((M_d[c])^TE(x_{d,i})/\tau)}{\sum_{c=1}^{C}exp((M_d[c])^TE(x_{d,i})/\tau)}
\end{align}
\replaced{where}{Where} $\tau$ is a temperature parameter.
\subsection{"Jury" Mechanism}
The domain-invariant features are directly related to the semantic features. We take $x^+$ which is \added{g}enerated by \deleted{the }data augmentation\added{,} as the semantically identical sample of input $x$. \replaced{We}{And we} introduce the "jury" mechanism mentioned in \cite{chen2021style} to ensure
the semantic similarity of each pair\added{,} x and $x^+$. To make the model evolve smoothly and maintain the consistency of representation over time, we construct an augmentation-related encoder\added{,} $E_k$\added{,} that updates parameters by momentum.
 
\textbf{Word Repetition. }
Following \cite{wu2021esimcse}, to change the semantics of the text as little as possible, we choose "word repetition" as our data augmentation method. We randomly repeat some words in a sentence. Given a text $x$ with words $w_1, ..., w_n$ in it, $x=\{w_1, w_2, ..., w_{N_{text}}\}$, where $N_{text}$ is the length of the text. We define the \replaced{maximum word}{word maximum} repetition rate \added{as} $r$. Then\added{,} \deleted{we have }the number of repeated words in a text, $k$ \added{is determined by random sampling within the range} $[0, maximum (2, int (r \times N_{text}))]$.\replaced{This parameter is used to expand the text length during word repetition. }{, which is the random sampling length. When the word repetition method is performed, this parameter will be used to expand the text length. }In this way, we \deleted{can }get the set $rep$ of all \deleted{the }repeated words, \deleted{where $rep$ is} obtained by uniform sampling. \replaced{If}{At this point, if} the words $w_1$ \added{and} $w_n$ are in the set $rep$, the text is converted to $x^+=\{w_1, w_1, w_2,..., w_n, w_n,... , w_{N_{text}}\}$.

\textbf{Momentum Update.}
We maintain an augmentation-related encoder $E_k$. Unlike \added{the} encoder $E_q$, which updates parameters \replaced{through}{by} back propagation, the encoder $E_k$ updates parameters \replaced{using}{with} momentum, as shown in Eq\added{u}.\textcolor{red}{\ref{eqn4}}:
\begin{align}\label{eqn4}
	\theta_k = \lambda \cdot \theta_k + (1-\lambda) \cdot \theta_q
\end{align}
\replaced{where}{Where} $\lambda$ is a momentum parameter; $\theta_q$ \added{and} $\theta_k$ \replaced{denote}{denotes} the parameters of encoder\added{s} $E_q$ \added{and} $E_k$\added{, respectively}. We believe \replaced{this}{such an} update method ensures the smoothness of \added{the} encoder $E_k$ update\added{s}, \replaced{reduces}{reducing} differences between \added{the} two encoders\added{,} and \replaced{maintains}{maintaining} temporal consistency.

\textbf{"Jury" Mechanism Related Loss. }
We construct domain-independent memories to store domain-invariant features. We \replaced{aim for}{hope} $x$ \replaced{to}{could} have a higher similarity with the semantically identical sample $x_{d,i}^+$ and a lower similarity with other samples. We build a domain-independent memory for each class to store the domain-invariant class feature\added{s}. We define $N_C$ domain-independent memories $Q=(Q_1,...,Q_c,...,Q_{N_C})$, where $c \in [1,N_C]$ \added{and} $N_C$ is the total \replaced{number of classes}{ class number}. Each domain-independent memory \replaced{is structured}{we build can be} as a queue of size $N_Q$:  $[q_c^1,...,q_c^j,...,q_c^{N_Q}]$, \added{where} $j \in [1,N_Q]$, \deleted{where $N_Q$ is the size of each domain-independent memory }and $q_c^j$ stores the class feature in position $j$ of the $Q_c$ domain-independent memory. We obtain the class features of each instance $x_{d,i}^+$ through \added{the} encoder $E_k$ and input them into the corresponding class's domain-independent memory \replaced{sequentially}{one by one}. \replaced{We}{Here, we} do not distinguish the domain of $x_{d,i}^+$, but store all $x_{d,i}^+$ sequentially into the corresponding class's domain-independent memory. We \replaced{place}{put} each newest feature at the end of the domain-independent memory and delete the oldest feature in the \deleted{domain-independent }memory.

All features in a domain-independent memory participate in \replaced{calculating}{the calculation of} the similarity \replaced{between}{of} $x_{d,i}$ and its augmented sample $x_{d,i}^+$. Both $x_{d,i}$ and $x_{d,i}^+$ \replaced{compute}{calculate} the cosine similarity with the domain-independent memory corresponding to the current sample’s class. We define $S_{d,i}=[s_1,...,s_j,...s_{N_Q}]$ as the similarity score between $x_{d,i}$ \replaced{and}{with} all class features stored in the corresponding class's domain-independent memory. We calculate the similarity score between $x_{d,i}$ and the corresponding domain-independent memory \replaced{using}{through} the softmax function, as shown in Eq\added{u}.\textcolor{red}{\ref{eqn5}}:
\begin{align}\label{eqn5}
	s_j =  \frac{exp((q_c^j)^TE_q(x_{d,i})/\tau)}{\sum_{j=1}^{N_Q}exp((q_c^j)^TE_q(x_{d,i})/\tau)}
\end{align}

Similarly, \replaced{we}{We} define $S_{d,i}^+=[s_1^+,...,s_j^+,...s_{N_Q}^+]$ as the similarity score between $x_{d,i}^+$ \replaced{and}{with} all class features stored in the corresponding class's domain-independent memory. The \replaced{method for calculating the}{way of} similarity score between $x_{d,i}^+$ and the domain-independent memory\replaced{ is}{, as} shown in Eq\added{u}.\textcolor{red}{\ref{eqn6}}:
\begin{align}\label{eqn6}
	s_j^+ =  \frac{exp((q_c^j)^TE_k(x_{d,i}^+)/\tau)}{\sum_{j=1}^{N_Q}exp((q_c^j)^TE_k(x_{d,i}^+)/\tau)}
\end{align}
Then\added{,} we penalize cross-entropy loss between \added{the} two similarity scores $s_j$ and $s_j^+$, as shown in Eq\added{u}.\textcolor{red}{\ref{eqn7}}:
\begin{align}\label{eqn7}
	L_{Jury} =  -\frac{1}{N_d}\sum_{i=0}^{N_d}s_j^+(x_{d,i}^+)log(s_j(x_{d,i}))
\end{align}
where $N_d$ is the \replaced{number of samples in the}{sample number of the} $d$-th source domain.

\subsection{Training Procedure}
We summarize the overall training process of the model in this section. Before training, we initialize the memory module and the \replaced{domain-independent memory}{dictionary} in the "jury" mechanism. During each iteration, \added{the} $D$ datasets \replaced{are}{is} randomly divided into one meta-test dataset and $D-1$ meta-train datasets. Then\added{,} the meta-train and meta-test stages \replaced{work together}{cooperate with each other} to optimize the model. 

\textbf{Meta-train. }
In the meta-train stage, \replaced{we first}{firstly, we} extract the same number of training samples $x_{d,i}$ from each meta-train domain. Then we input these samples into encoder $E_q$ to extract features $E_q(x_{d,i})$. \replaced{ The features are subsequently fed into classifier $C$ to calculate classification losses and memory-based similarity loss with the slots in the memory module.}{These features are then input into the classifier $C$ to calculate classification losses and calculated memory-based similarity loss with slots in the memory module.} Besides, we input augmented samples $x_{d,i}^+$ into encoder $E_k$ to extract augmented features $E_k(x_{d,i}^+)$. \replaced{The}{Then the} "jury" mechanism\added{-}related loss is calculated with features $E_q(x_{d,i})$ and augmented features $E_k(x_{d,i}^+)$. Finally, we update the \replaced{domain-independent memory}{dictionary} built for the "jury" mechanism. \replaced{The meta-train}{Meta-train} loss \replaced{comprises the}{is composed of} meta-train classification loss, memory-based similarity loss\added{,} and the "jury" mechanism\added{-}related loss. The formula is shown \replaced{in}{as Equation} Equ.\textcolor{red}{\ref{eqn8}}:
\begin{align}\label{eqn8}
\begin{split}
	L_{mtr}^d=L_{C}(X_d;\theta_q;\theta_C)+L_{Mem}(X_d;M_d;\theta_q)+\\L_{Jury}(X_d;Q;\theta_k)
\end{split}
\end{align}
where $\theta_q$, $\theta_C$, $\theta_k$ denote parameters of the encoder $E_q$, the classifier $C$, the encoder $E_k$ \added{respectively}; $X_d$ and $M_d$ denote \added{the} samples and \added{the} memory module of domain $d$; $Q$ denotes the \replaced{domain-independent memory}{dictionary}.

The total meta-train loss is the averaged \replaced{of the losses from all meta-train domains}{number of all meta-train domains}:
\begin{align}\label{eqn9}
	L_{mtr}=\frac{1}{D-1}\sum_{d=1}^{D-1}L_{mtr}^d
\end{align}

\textbf{Meta-test. }
In the meta-test stage, \replaced{we first}{firstly, we} copy the encoder $E_q$ and classifier $C$. Then\added{,} we update them with \replaced{the meta-train loss}{loss with} $L_{mtr}$. \replaced{We}{And we} update \added{the} encoder $E_k$ with parameters $\theta'_q$, where $\theta'_q$ \replaced{represents}{is} the updated parameters of $E_q$. The meta-test loss is calculated in the same way as the meta-train loss. \replaced{It is}{Meta-test loss is also} composed of the meta-test classification loss, the memory-based similarity loss, and the "jury" mechanism\added{-}related loss. The formula is shown \replaced{in}{as Equation} Equ.\textcolor{red}{\ref{eqn9}}:
\begin{align}\label{eqn10}
\begin{split}
	L_{mte}=L_{C}(X_T;\theta'_q;\theta'_C)+L_{Mem}(X_T;M_T;\theta'_q)+\\L_{Jury}(X_T;Q;\theta'_k)
\end{split}
\end{align}
where $\theta'_q$, $\theta'_C$, \added{and} $\theta'_k$ denote \added{the} optimized parameters of the encoder $E_q$, the classifier $C$, \added{and} the encoder $E_k$\added{, respectively}; $X_d$ and $M_d$ denote \added{the} samples and \added{the} memory module of d domain; \replaced{$Q$ denotes the domain-independent memory}{dictionary}.

\textbf{Meta Optimization. }
Finally, we \replaced{optimize}{optimized} the model as \replaced{shown in}{the Equation} Equ.\textcolor{red}{\ref{eqn1}}. \replaced{Our}{And our} training procedure is \replaced{detailed in}{as the algorithm} Alg.\textcolor{red}{\ref{alg1}}:
\begin{algorithm}[h]  
\caption{Training Procedure of MMJM} 
\label{alg1}
\KwIn{$D$ source domains $D_{S}=\{D_{S}^d\}_{d=1}^D$.} 
\SetKwInOut{Output}{Initialize}
\Output{$E_q$ parameterized by $\theta_q$; $C$ parameterized by $\theta_C$;\\
$E_k$ parameterized by $\theta_k$; Batch size $B$\\
Inner and outer loop learning rates $\alpha$ and $\beta$;\\ }

\textbf{Memory Module Initialization:}\\
\ForEach{d}{
Extract all the samples' features in domain $d$;\\
Average features according to classes $C$;\\
Initialize memory slots $M_d[C]$ with average values.\\
}
\textbf{\replaced{Domain-invariant memory}{Dictionary} Initialization:}\\
Random initialize \replaced{the domain-invariant memory}{dictionaries} $Q$.\\
\ForEach{iter}{
Randomly split $D$ into $D_{mtr}$ and $D_{mte}$;\\
\textbf{Meta-train:}\\
Sample batch $B$ from each domain $X_d = \{x_i\}_{i=1}^B$\\
Compute meta-train loss $L_{mtr}$ with \deleted{equation }Equ.\textcolor{red}{\ref{eqn9}};\\
\textbf{Meta-Test:}\\
Sample batch $B$ from meta-test domain $X_T = \{x_i\}_{i=1}^B$;\\
Copy the encoder $E_q$ and classifer $C$ and update $\theta_{q,C}$ :
$\theta'_{q,C}\leftarrow Adam(\nabla_{\theta}L_{mtr},\theta_{q,C},\alpha)$;\\

Compute meta-test loss $L_{mte}$ with \deleted{equation }Equ.\textcolor{red}{\ref{eqn10}};\\
Update \replaced{the domain-invariant memory}{dictionaries} $Q$;\\
Update $\theta_{k}$ with the \deleted{equation }Equ.\textcolor{red}{\ref{eqn4}}\\
Update \replaced{all domain-specific memory}{dictionaries} $M$;\\
\textbf{Meta Optimization:}\\

Compute gradient: $g\leftarrow Adam(\nabla_{\theta_{q,C}}(L_{mtr}(\theta_{q,C})+L_{mte}(\theta_{q,C}'))$\\
Update $\theta_{q,C}$ :
    $\theta_{q,C} \leftarrow Adam(g,\theta_{q,C},\beta)$}
    
\end{algorithm}

\section{Experimental Setup}
\subsection{Experimental Data \added{and Evaluation Metrics}}
To verify the effectiveness of this method, we conduct experiments on two multi-source text classification datasets: the Amazon product review dataset \cite{blitzer2007biographies} for the multi-source sentiment analysis, and the multi-source rumor detection dataset \cite{zubiaga2016analysing}. \added{The Amazon product review dataset contains 8,000 reviews, evenly distributed across four domains: Books (B), DVDs (D), Kitchens(K), and Electronics (E). Each domain has 1,000 positive reviews and 1,000 negative reviews. The multi-source rumor detection dataset consists of 5,802 annotated tweets from five different events: Charlie Hebdo (CH), Ferguson (F), Germanwings (GW), Ottawa Shooting (OS), and Sydney Siege (SS), labeled as rumors or non-rumors (1,972 rumors, 3,830 non-rumors). }

\added{In the experiments, for each dataset, we alternately select one domain as the test set while using the remaining domains as the training set. For evaluation, we use the average accuracy for multi-source sentiment analysis and the average F1 score for multi-source rumor detection, calculated by averaging the results from experiments where each domain is used as the test set.}

\subsection{Baselines}
We compare \added{our method} with \replaced{several}{some} state-of-the-art approaches. 
\added{However, previous methods are mostly based on the Domain Adaptation setting. For fairness, we employ these methods under the Domain Generalization setting.}

\added{
\textbf{Basic} fine-tunes a model on labeled data from source domains and directly tests it on the target domain. \textbf{Gen} refers to \cite{li2018s}, which proposes a model composed of several domain-specific CNNs to compute private representations and a shared CNN to compute shared representations, coupled with adversarial training. The MoE model consists of dedicated models for each source domain and a global model trained on labeled data from all source domains. During inference, the ensemble predictions of all models are aggregated. In this context, MoE~\cite{wright2020transformer} refers to a MoE model without a pretrained model, while \textbf{MoE-Avg}~\cite{wright2020transformer} refers to a MoE model with a pretrained model. \textbf{PCL} refers to a proxy-based contrastive learning method~\cite{yao2022pcl}, where the traditional sample-to-sample mechanism is replaced by the proxy-to-sample mechanism. \textbf{Intra}~\cite{wen2016discriminative,ye2020feature} refers to center loss, which minimizes the distance between each example and its class center. \textbf{Adv} refers to the well-studied domain adversarial adaptation method~\cite{ganin2016domain}, which reverses the gradient calculated by the domain classifier. \textbf{Agr-Sum}~\cite{mansilla2021domain} refers to two gradient agreement strategies based on gradient surgery to reduce the effect of conflicting gradients during domain generalization.
}


\begin{table*}[h] 
    \centering
    \caption{Experiments Results Compared with State-of-the-art Approaches.}
    \begin{tabular}{c|ccccc|cccccc}
        \toprule
        Method & D & B & E & K  & Avg Acc & CH & F & GW & OS &S  & Avg F1         \\
        \midrule
        Gen & 77.9&77.1 & 80.9& 80.9 & 79.20&-&-&-&-&-&-\\
        MoE & 87.7&87.9 & 89.5& 90.5& 88.90&-&-&-&-&-&-\\
        \midrule
        DistilBert\\
        \midrule
        MoE-Avg &88.9&90.0&90.6&90.4&89.98 &67.9&45.4&74.5&62.6&64.7&63.02\\
        SCL&\textbf{90.1}&90.0&90.3&\textbf{90.8}&90.30&\textbf{68.1}&44.5&75.4&66.5&65.2&63.94\\
        Basic&89.1&89.8&90.1&89.3&89.58&66.1&44.7&71.9&61.0&63.3&61.40\\
        MLDG&89.5&90.3&\textbf{90.8}&90.7&90.33&66.1&52&\textbf{79.8}&69.9&63.1&66.18 \\
        PCL &89.2&89.8&90.3&90.5&89.95&60.4&50.1&75.7&70.9&64.5&64.32\\

        Intra&88.5&89.8&90.1&89.2&89.40&64.1&42.9&70.8&61.8&62.4&60.40 \\
        Adv&88.4&89.0&89.6&90.0&89.25&64.8&42.2&65.9&61.4&62.8&59.42\\
        Agr-Sum&88.8&89.1&90&90.3&89.55&67.5&52.0&76.9&69.0&64.3&65.94\\
        \textbf{MMJM}&89.8&\textbf{90.5}&\textbf{90.8}&\textbf{90.8}&\textbf{90.48}&66.3&\textbf{52.3}&77.1&\textbf{73.4}&\textbf{72.7}&\textbf{68.36}	\\
        \midrule
        Bert\\
        \midrule
        MoE-Avg&90.4&91.4&91.5&92.3&91.4&67.7&46.7&\textbf{80.8}&53.7&60.5&61.88\\
        Basic &90.5&91.2&92.2&91.9&91.45&66.6&46.0&73.7&69.2&61.8&63.46\\
        MLDG &91.1&91.9&91.8&92.5&91.83&67.8&50.3&77.3&71.3&60.5&65.44\\
        PCL &89.5&91.4&\textbf{92.4}&91.8&91.28&64.1&\textbf{53.9}&70.1&70.9&66.2&65.04\\
        Intra&90.4&91.2&91.3&92.1&91.25&63.4&47.9&71.0&67.6&57.7&61.52\\
        Adv &91.1&91.2&91.1&92.0&91.35&66.1&44.4& 69.0&70.3&62.2&62.40\\
        Agr-Sum&90.2&91.1&91.2&91.8&91.08&67.9&53.4&77.3&70.5&57.6&65.34\\
        \textbf{MMJM}&\textbf{91.2}&\textbf{91.7}&91.9&\textbf{92.7}&\textbf{91.88}&\textbf{68.8}&52.6&78.3&\textbf{73.0}&\textbf{70.6}&\textbf{68.66}\\
        \bottomrule
    \end{tabular}
    \label{tab:comparison}
    \vspace{-10pt}
\end{table*}

\subsection{Implement Details}
For all experiments, due to GPU memory constraints, we adapt Distil-Bert-base-uncased \cite{sanh2019distilbert} and Bert-base-uncased \cite{devlin2018bert} pretrained models as our encoders. The training batch size is set to 8, and we \replaced{train}{trained} 15 epochs. We set the token number of samples to 512. To optimize our model, we use \added{the} Adam optimizer with a weight decay of $5\times10^{-4}$ to optimize the encoder $E_q$. The inner loop learning rate $\alpha$ and \added{the} outer loop learning rate $\beta$ \replaced{start at}{are started as} $1 \times 10^{-6}$ and \replaced{are increased}{heated up} to $1 \times 10^{-5}$ in the first epoch. For the memory module, the momentum coefficient $m$ is set to 0.2 and the temperature factor $\tau$ is set to 0.05. For the "jury" mechanism, the momentum coefficient $\lambda$ is set to 0.999, and the size of the domain-independent memory is set to $64 \times 768$. 

\subsection{Experimental Results}
The experimental results of text classification are shown in Tab.\textcolor{red}{\ref{tab:comparison}}. First, in the multi-source sentiment analysis and rumor detection, our method achieves the best performance, \replaced{demonstrating its effectiveness}{this can show that our method is effective} for DG in the text classification. Second, compared to the meta-learning baseline MLDG and the contrastive learning baseline SCL, our method \replaced{achieves}{achieve} higher accuracy. This success highlights the advantage of capturing both domain-invariant and domain-specific features. Third, our method does not reach peak performance in every domain. We attribute this to the ambiguous characteristics of data in some areas. Nonetheless, our strong performance across domains, as evidenced by average scores, highlights the method's adaptability in diverse settings.

\subsection{Ablation studies}
To further analyze the effectiveness of each part of our model, we conduct ablation studies. The results are shown in Tab.\textcolor{red}{\ref{tab:ablation}}, where "Meta" \replaced{indicates}{:} training with the meta-learning framework, "Mem" \replaced{indicates}{:} training with the memory module, "Jury" \replaced{indicates}{:} training with the "jury" mechanism, "SA" \replaced{refers}{:} the multi-source sentiment Analysis, and "RD" \replaced{refers}{:} the multi-source rumor detection. All ablation studies are based on distil-Bert. 

\begin{table}[h] 
    \centering
    \caption{Ablation Experiments}
    \begin{tabular}{c|c|c|c|c|c|c}
        \toprule
        Meta& Mem&Jury &  SA Avg Acc   & RD Avg F1 &RAM usage&GPU usage        \\
        \midrule
       $\times$&   $\times$ &$\times$& 89.58&61.40&2711.71 MB&1058.31 MB\\
        $\checkmark $ & $\times$&$\times$& \textbf{90.6}&69.00&2871.52 MB&1816.93 MB\\
        $\times$&$\times$ & $\checkmark $& 90.28&66.96&2764.84 MB&1312.36 MB\\
        $\checkmark $&$\checkmark $&$\times$& 90.38&68.30&2941.57 MB& 1817.65 MB\\
        $\checkmark $&$\times $&$\checkmark$& 90.33&\textbf{69.32}&2894.88 MB&2072.82 MB\\
        $\checkmark $&$\checkmark $&$\checkmark $& 90.43&68.36&\textbf{3031.73 MB}&\textbf{2073.60 MB}\\
        		
        \bottomrule
    \end{tabular}
    \label{tab:ablation}
\vspace{-10pt}
\end{table}

\textbf{Effectiveness of Meta-Learning.}
We conduct the ablation study to investigate the proposed meta-learning strategy\deleted{'s effectiveness}. The model trained with the proposed meta-learning strategy could improve the results. For sentiment analysis, the model trained with a meta-learning strategy \replaced{increases}{could increase} the basic baseline by $1.02\%$ \deleted{improvement} in the average classification accuracy. For rumor detection, the average F1 score increases by $7.6\%$. We believe that the implementation of meta-train and meta-test processes within the meta-learning framework could help the model adapt to the training of multi-source domains and learn domain-related features. This approach potentially reduces the risk of the model overfitting to domain biases, thereby \replaced{enhancing}{bolstering} the model's performance in encountering unseen domains. \added{However, the introduction of the meta-learning approach consumes substantial computational resources and memory. We believe this is due to the characteristics of the meta-learning method, which combines data from multiple source domains and updates parameters in two stages.}

\textbf{Effectiveness of the "Jury" Mechanism.} The "jury" mechanism related loss could increase the classification accuracy. For sentiment analysis, compared with the Distil-Bert baseline, the model trained with the "jury" mechanism could \replaced{achieve an average accuracy increase of}{increase the average accuracy} $0.7\%$. For rumor detection, the average F1 score increases \added{by} $5.56\%$. We believe that this is because the model is trained with a large number of data-augmented examples, \replaced{which}{this method} could help the model learn to reduce the domain divergence between different domains. What's more, it's obvious that the average F1 score increases \added{by} $0.32\%$ for rumor detection, but the average accuracy reduces \added{by} $0.27\%$ for sentiment analysis when \replaced{combining}{we combined} the meta-learning method and the "jury" mechanism. We think that this observation may be attributed to the more apparent similarities among data from different domains in sentiment analysis. \added{Additionally, we find the 'jury' mechanism demands relatively fewer computational resources and memory. It is a low-cost method that can enhance the model's generalization capabilities.}

\textbf{Effectiveness of Memory Module.}
\replaced{Contrary to expectations,}{It is different from what was expected} the meta-learning framework with the memory module can't further increase the accuracy of classification compared \replaced{to using meta-learning alone}{with meta-learning}. However, compared to the model without the memory module, MMJM could increase the average sentiment analysis accuracy \added{by} $1\%$. We believe that the memory module is helpful when combined with the domain-invariant features. \added{Additionally, the memory module requires minimal computational resources and relatively low memory. We believe that the introduction of the memory module, despite its low cost, can significantly enhance the generalization capability of the entire framework. Thus, the inclusion of the memory module is highly worthwhile.}

\subsection{Visualization}

\begin{figure} [htbp]
	\centering
	\subfloat[Distil-Bert 
	\label{tsnebasic}]{
		\includegraphics[scale=0.07]{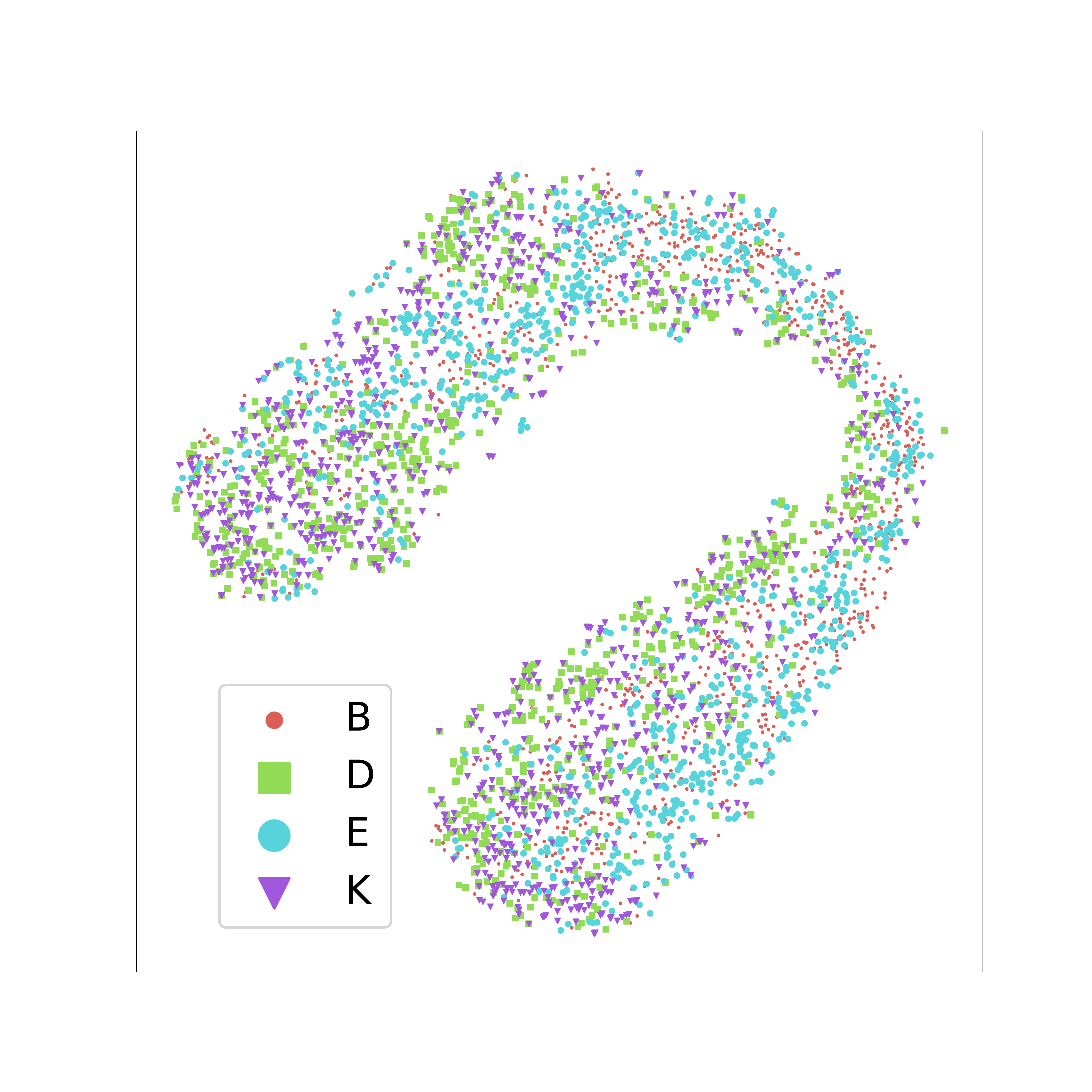}}
	\subfloat[PCL 
             \label{tsnepcl}]{
		\includegraphics[scale=0.07]{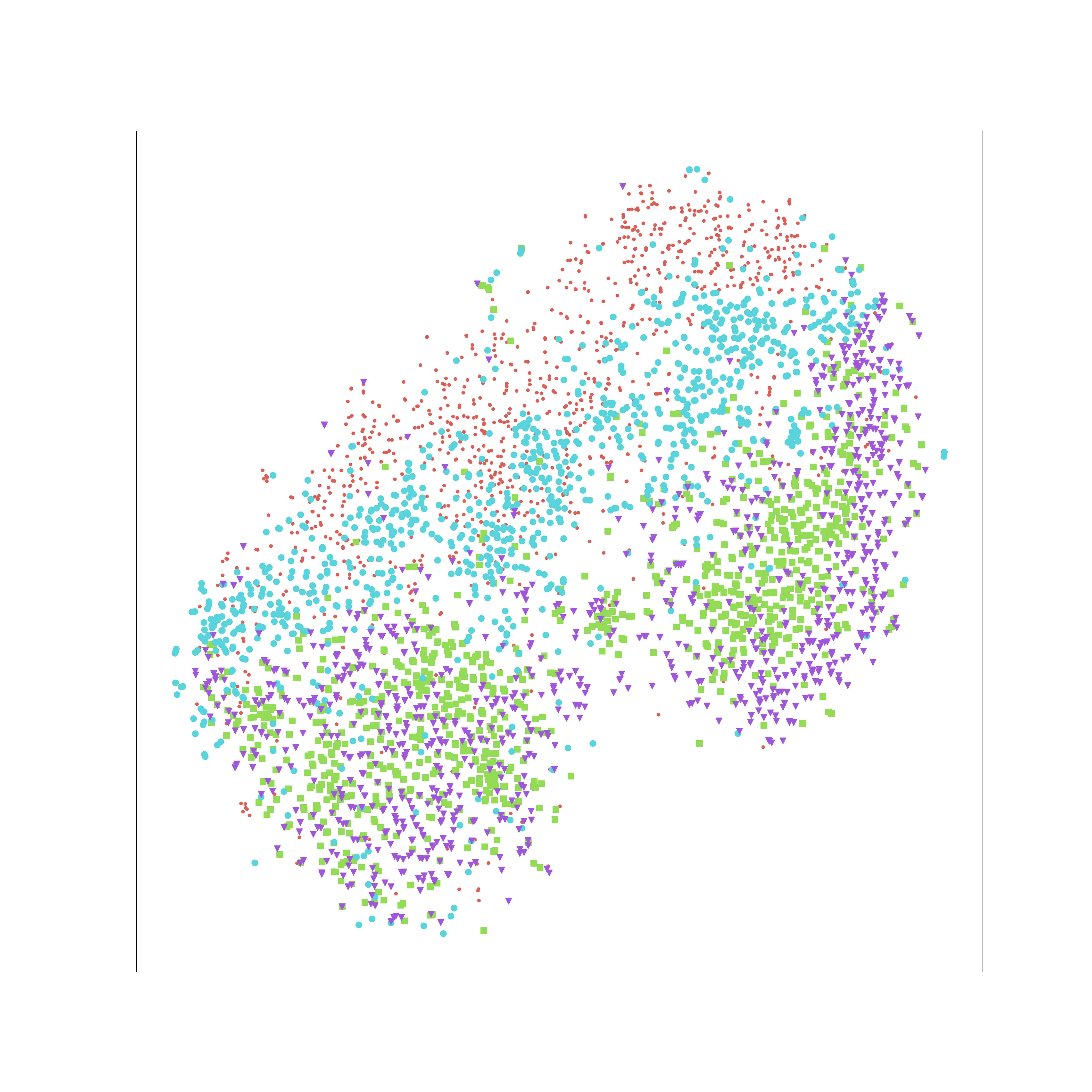} }
    \subfloat[MMJM 
	\label{tsnemodel}]{
		\includegraphics[scale=0.07]{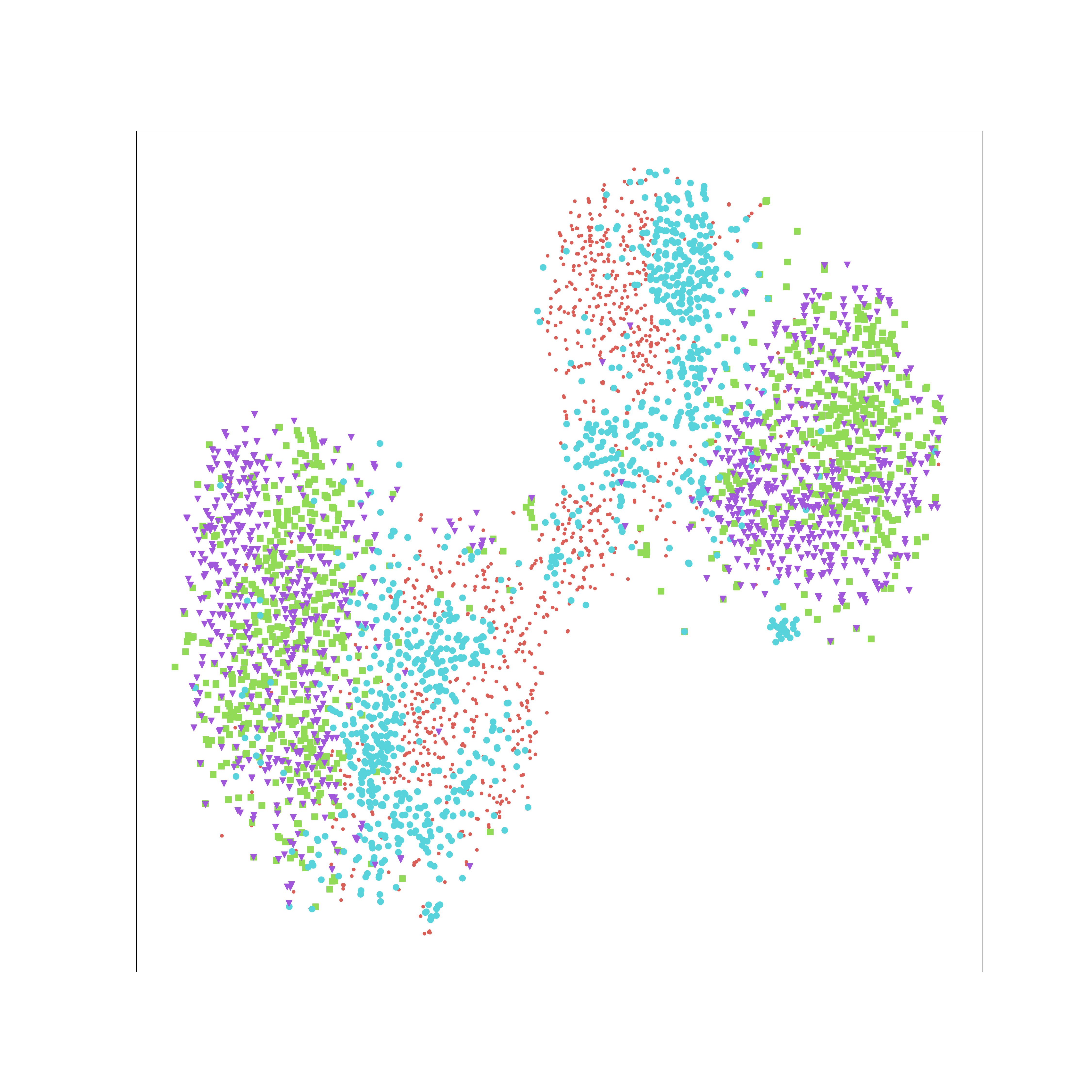}}
	\caption{t-SNE visualization of the embeddings from Distil-Bert. We choose models with source domains D,K,E and target domain B and sample 1,000 examples for each domain.}
	\label{TSNE} 
\end{figure}

To better understand the effectiveness of our method, \replaced{we}{We also} provide the t-SNE visualizations \cite{van2008visualizing} of our MMJM and some baselines to intuitively \replaced{assess}{learn} the performance of our model on domain discrepancy, as illustrated in Fig.\textcolor{red}{\ref{TSNE}}. We observe that features from the source and target domains of MMJM are much more compact, which indicates our framework can learn more domain-specific and domain-invariant features.
\vspace{-15pt}
\vspace{-15pt}

\subsection{Comparison with Large Language Model}
\added{
To ensure the comprehensiveness of our experiments, we also explored the performance of large language models on the datasets. The experimental results are shown in Tab.\ref{tab:llm}. It can be seen that on the sentiment classification dataset, ChatGPT\footnote{gpt-3.5-turbo} performs well, with an average classification accuracy $2.1\%$ higher than MMJM. This indicates that large language models have strong sentiment perception abilities and can achieve good results across different domains, demonstrating strong generalization capabilities. However, on the rumor dataset, its performance is poor, which we believe is due to the lack of relevant knowledge about rumor detection in the language model. Therefore, we believe our model has research value, as it uses only $1\%$ of the parameters of ChatGPT yet achieves good classification performance, significantly saving training resources and time costs for training large models. Additionally, it can still achieve good results through training on specific topics.
}
\begin{table*}[h] 
    \centering
    \caption{Experiments Results Compared with ChatGPT.}
    \begin{tabular}{c|ccccc|cccccc}
        \toprule
        Method & D & B & E & K  & Avg Acc & CH & F & GW & OS &S  & Avg F1         \\
        \midrule
        ChatGPT & \textbf{93.1} & \textbf{93.2}&\textbf{94.4}&\textbf{95.2}&\textbf{93.98}&40.9&36.5&52.0&39.3&48.9&43.52\\
        MMJM&91.2&91.7&91.9&92.7&91.88&\textbf{68.8}&\textbf{52.6}&\textbf{78.3}&\textbf{73.0}&\textbf{70.6}&\textbf{68.66}\\
        \bottomrule
    \end{tabular}
    \label{tab:llm}
    \vspace{-5pt}
\end{table*}
\subsection{Case Study}

We present a case study to intuitively understand our framework mechanism, as shown in Tab.\textcolor{red}{\ref{tab:case}}. The encoder for both \replaced{models}{of them} is distil-Bert. P and N respectively denote the positive and negative predictions. The symbol \ding{56} indicates \replaced{a}{the} wrong prediction. Cases from four domains B, D, E, and K. The second sentence shows that both the basic baseline and our model MMJM can make the correct prediction. We believe that is because the sentence is relatively simple \replaced{making it easier to identify key information such as}{so it seems not too hard to catch the key information like} "books" and "stupid" \deleted{in this sentence}. However, the performances of the two models in the third sentence are different. That demonstrates that our MMJM model can address complicated and informal sentences, but the basic model focuses on the sentiment words "kind" and "clear". What's more, we find both our model and the basic model would make the wrong answer when the class-related information or sentiment expression is unclear, like the first and the fourth sentences. Specifically, the fourth sentence \replaced{contains}{has} class-related words \added{like} "rip" and "hard", which leads to the mistake, \replaced{while}{of this sentence and} the first sentence has an unclear sentiment expression. ALL in all, our method makes the true prediction most of the time, so we make the conclusion that our model MMJM \replaced{can capture}{could catch} the domain-related and class-related information, \replaced{though}{but} there is still room for improvement.

\begin{table}[h!t]

\center
\caption{Case study between the proposed framework and the basic model. }
\begin{tabular}{p{260pt}<{\raggedright}p{40pt}<{\centering}p{40pt}<{\centering}}
  \toprule
  Text& Basic&MMJM \\
  \midrule
  I saw the scene, where they have lissa chained to the pool table and gagged in
the basement. I didn't understand most of the movie. I bet kim possible, Ron
Stoppabl, and rufus can deal with them.(D)&   N(\ding{56})&N(\ding{56})\\
  \midrule
  I really feel stupid about ordering this book. Why did I do it?. The story is that
stacey is moving back to new york.(B)    & N&N \\
\midrule
Very clear image! Probably the best that I saw on this kind devices! No software
availble. Device is useless if you do not have windows media os. Absolutely no
linux support. Good hardware... but useless... you have to buy software for ~\$70-
120 to be able use it...(E)& P(\ding{56})&N\\
\midrule
It really does make a difference when some of the chlorine is filtered out of your
coffee (and even more so in your bourbon) water, but these filters are a rip off
price-wise. It's hard to believe braun wouldn't make money selling them at a
third of the price.(K)& (\ding{56})&N(\ding{56})\\
  \bottomrule
\end{tabular}
\label{tab:case}
\end{table}

\section{Conclusion}
In this paper, we propose a multi-source meta-learning framework for DG in text classification. The meta-learning strategy simulates how the model generalizes to an unseen domain. Additionally, we incorporate a memory-based module and the "jury" mechanism to extract domain-invariant features and domain-specific features, further enhancing the model's performance.

%
%
%
%

\end{document}